\documentclass[letterpaper, 10 pt, conference]{ieeeconf}  

\IEEEoverridecommandlockouts                              

\usepackage{amsmath} 
\usepackage{amssymb}  
\usepackage{todonotes}
\let\labelindent\relax 
\usepackage{enumitem}
\usepackage{mathtools}
\usepackage{rotating}
\usepackage{multirow}
\usepackage{siunitx}
\usepackage{caption}
\usepackage{subcaption}
\usepackage{float}

\usepackage{booktabs} 

\usepackage{algorithmic} 
\usepackage{algorithm2e}
\graphicspath{{graphics/}}

\usepackage{tabto} 

\title{\LARGE \bf
LSM: A Comprehensive Metric for Assessing the Safety of Lane Detection Systems in Autonomous Driving
}

\author{Jörg Gamerdinger$^{1}$, Sven Teufel$^{1}$, Stephan Amann$^{1}$, Georg Volk$^{1}$, and Oliver Bringmann$^{1}$
\thanks{$^{1}$University of T\"ubingen, Faculty of Science, Department of Computer Science, Embedded Systems Group 
\tt\small {\{joerg.gamerdinger, sven.teufel, stephan.amann, georg.volk, oliver.bringmann\} @uni-tuebingen.de}}
}%

\begin{document}
\maketitle
\thispagestyle{empty}
\pagestyle{empty}

\begin{abstract}
Comprehensive perception of the vehicle's environment and correct interpretation of the environment are crucial for the safe operation of autonomous vehicles. The perception of surrounding objects is the main component for further tasks such as trajectory planning. However, safe trajectory planning requires not only object detection, but also the detection of drivable areas and lane corridors. While first approaches consider an advanced safety evaluation of object detection, the evaluation of lane detection still lacks sufficient safety metrics.
Similar to the safety metrics for object detection, additional factors such as the semantics of the scene with road type and road width, the detection range as well as the potential causes of missing detections, incorporated by vehicle speed, should be considered for the evaluation of lane detection. Therefore, we propose the Lane Safety Metric (LSM), which takes these factors into account and allows to evaluate the safety of lane detection systems by determining an easily interpretable safety score. We evaluate our offline safety metric on various virtual scenarios using different lane detection approaches and compare it with state-of-the-art performance metrics.

\end{abstract}


\section{INTRODUCTION}
\label{sec:intro}
Human error is responsible for \SI{90}{\percent} of fatal traffic accidents in the European Union~\cite{EuropeanUnion2019}. Autonomous driving is predicted to increase road safety and improve traffic flow by eliminating human error. For this, automated vehicles must overcome a number of challenges before they are ready to enter the market. Accurate and complete perception of the environment is one of the most important issues. Unfortunately, perception is not always accurate, which increases the risk of fatal car accidents involving autonomous vehicles~\cite{uber-accident}.

However, to achieve safe systems, safety must be evaluated, not just performance. Therefore, an appropriate metric is required to evaluate the system under test. While commonly used performance metrics do not take into account the semantics of the scene, such as the velocity and the vulnerability of the surrounding road users, improved metrics take these factors into account~\cite{volk2020safety}. 
Currently, safety evaluation focuses on the environmental perception of objects; however, also lanes must be correctly perceived to plan valid and safe maneuvers. Therefore, it is necessary to develop new metrics for evaluating the safety of lane detection systems in order to achieve safe autonomous driving.
\begin{figure}[t]
    \centering
    \includegraphics[width=.90\linewidth, page=4, trim= 10cm 4cm 10cm 4.2cm, clip]{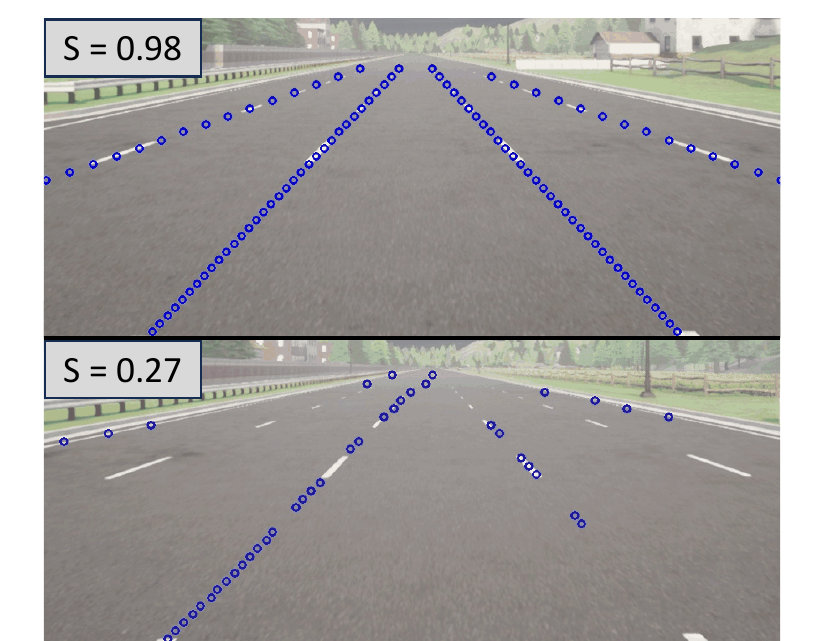}
    \caption{Example for safe detection (top) and unsafe detection (bottom).} 
    \label{fig:example}
    \vspace*{-3mm}
\end{figure}

The main contributions of this work are:
\begin{itemize}
    \item We demonstrate the limitations of performance metrics for the evaluation of lane detection systems
    \item We propose a novel offline metric to evaluate safety of lane and drivable corridor detection 
\end{itemize}
 In Sec.~\ref{sec:related_work}, we briefly discuss work related to safety evaluation and lane detection. The structure and relevant factors of our metric are explained in Sec.~\ref{sec:method}. Afterwards, we present the results of our proposed metric based on three different virtual scenarios in comparison to state-of-the-art performance metrics in Sec.~\ref{sec:results}. Finally, in Sec.~\ref{sec:conclusion} we conclude our work and give an outlook to possible extensions and further research.

\section{RELATED WORK}
\label{sec:related_work}

Dupuis et al.~\cite{dupuis2010opendrive} demonstrated the description of lanes as lines, arcs, clothoids, and polynomial functions. Also splines are capable to represent lanes with varying curvatures and layouts very well as shown in~\cite{wang2000lane,zhao2012novel}.
A network of multiple lanes forms a map that may also include additional information, such as the surrounding area or the placement of buildings. 
A widely used map representation is the OpenDrive format by Dupuis et al.~\cite{dupuis2010opendrive}. OpenDrive allows to describe the centerline with mathematical primitives and single lanes as relation to this centerline. Also infrastructural elements such as traffic signs or signals can be described by this format.
A simple way to represent lanes as lists of points for both lane boundaries is proposed in the Lanelet framework by Bender et al.~\cite{lanelet}. Moreover, by the extensions in Lanelet2~\cite{lanelet2} the framework provides multiple functions for handling and querying maps as well as additional modules for routing or the representation of traffic rules. 


State-of-the-art lane detection benchmarks~\cite{TuSimple, CULane, Fritsch2013ITSC, LLamas} use the pixel-based performance metrics precision $P$ and recall $R$ as defined in Eq.~\eqref{eq:prec}: 
\begin{equation}
    \label{eq:prec}
    P = \frac{TP}{TP+FP},\qquad R = \frac{TP}{TP+FN}, 
\end{equation}

where true positive (TP) describes a correctly detected pixel within a threshold of about 10 to 30 pixels, false positive (FP) stands for a pixel which is erroneously classified as lane pixel. False negative (FN) describes a pixel of the lane which is not classified as lane.
Resulting from these metrics the $F1$-score as defined in Eq.~\eqref{eq:f1} can be derived. F1 is conducted to incorporate the advantages of both metrics.
\begin{equation}
    \label{eq:f1}
    F1 = 2\cdot\frac{P\cdot R}{P+R}
\end{equation}

An optimization is presented by Fritsch et al.~\cite{Fritsch2013ITSC}. They propose a behavior-based method where the driving corridor with the highest probability based on the detected boundaries is evaluated using precision and recall. A metric evaluating the highest lateral deviation in meter between vehicle center and the centerline of the lane was presented by Sato et al.~\cite{sato2022towards}. The metric can be used for a complete scenario using the end-to-end mode or per frame. For some scenarios the result in meters can be more meaningful but still cannot validate the safety as information about the vehicle and the scene are neglected.

However, Volk et al.~\cite{volk2020safety} have shown that performance metrics do not allow to make a statement about the safety of the perception. They present a novel metric to evaluate the environmental perception of objects by considering the semantics of the scene by including different factors such as velocity and the severity of a potential collision. Since, the metric is constructed for object perception it cannot be used for the evaluation of lane detection systems.

For the safety evaluation the criticality of an object to be detected is an important factor. Criticality describes how crucial the perception of an object is or how relevant it is to incorporate the object in the motion planning as neglecting the object could lead to safety-critical situations. An often used criticality metric is the distance to an object. A well-known approach defining such safety distances to rate criticality is the Responsibility-Sensitive Safety (RSS) model by Shalev-Shwartz et al.~\cite{RSS}. However, the model aims to provide a mathematical formalization of traffic behavior for interactions between vehicles which allows no application for lane detection. An approach defining risk zones is presented by Ca\~{n}as et al.~\cite{dynamicRiskAssessment}. Risk zones are defined based on interior and exterior perception in combination with the vehicle dynamics; however, the approach was only designed for parking scenarios with velocities up to \SI{5}{\kilo\metre\per\hour}. A comprehensive survey about criticality metrics for perception and motion planning is presented by Westhofen et al.~\cite{westhofen2023criticality}; however, metrics to evaluate the safety are not included.

Vehicle-local lane detection is investigated in multiple works as shown by Mamun et al.~\cite{mamun2022comprehensive}. Early image-filtering-based approaches using edge detection and Hough transformation were presented by Wang et al.~\cite{wang2000lane} and Aly~\cite{aly2008real}.
Qin et al.~\cite{qin2020ultra} proposed a neural network based ultra fast lane detection approach which shows a speedup of up to 43 times compared to other state-of-the-art lane detectors while maintaining an accuracy of about \SI{96}{\percent} on TUSimple dataset~\cite{TuSimple}. An anchor-based approach was presented by Tabelini et al.~\cite{tabelini2021cvpr}. This approach is capable of achieving similar results as~\cite{qin2020ultra}, but shows a reduced false positive rate.
However, local perception may not always satisfy sensing range requirements. To overcome this, Gamerdinger et al.~\cite{gamerdinger2023cold} proposed a collective lane detection to increase the safety. The approach fuses perceived lanes from different vehicles to achieve a more accurate lateral detection and a significant increase in the detection range.
\begin{figure*}[t]
    \centering
    \includegraphics[width=0.9\linewidth, page=2, trim= 1.5cm 6.5cm 1.5cm 0cm, clip]{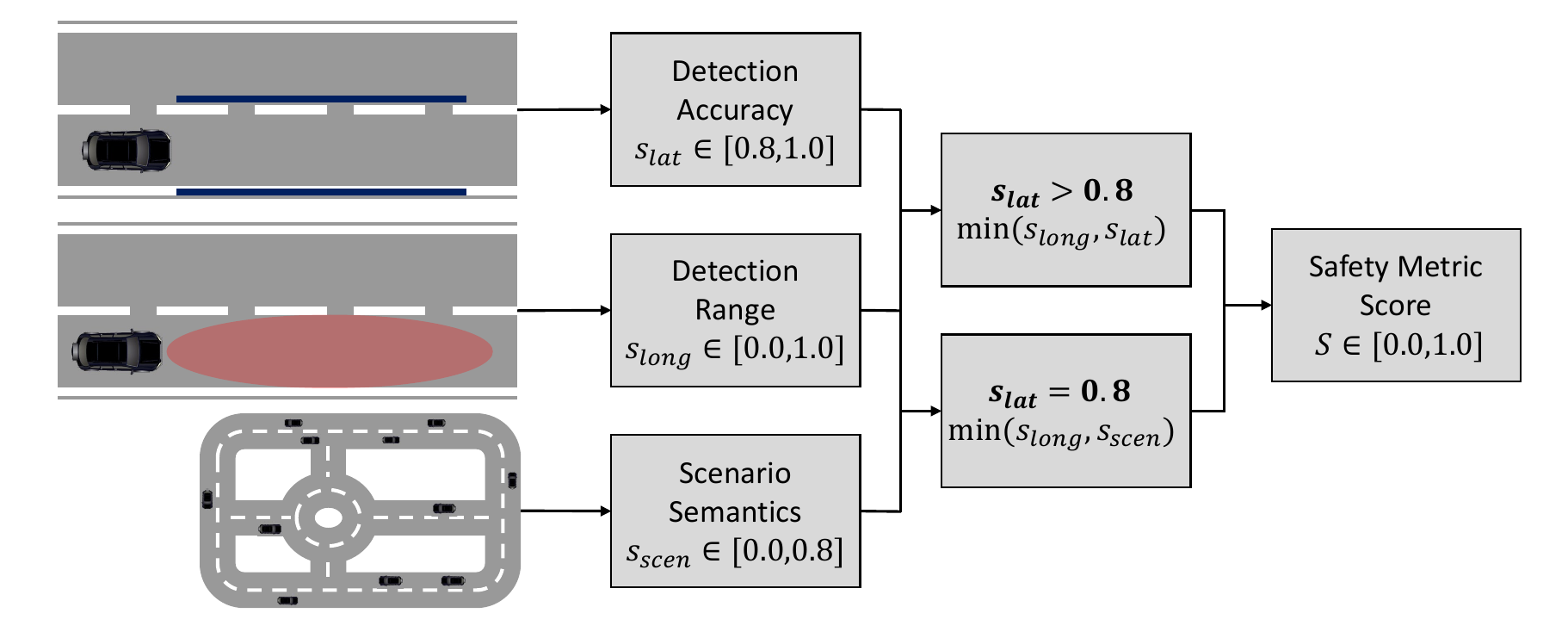}
    \caption{Schematic overview of the proposed LSM including their intermediate metrics and their composition to a single safety score $S$.} 
    \label{fig:overview}
\end{figure*}
\section{Lane Safety Metric}
\label{sec:method}

The goal of this work is to develop a metric that allows for a comprehensive evaluation of the safety and not just the performance of lane detection systems. The metric has the requirement to allow an easy evaluation and comparison between different lane detection systems under various conditions. To achieve this, the result of the metric is a single value $\in [0,1]$. Additionally, the metric shall be adaptable to different controller models and scenarios.
Our proposed metric is based on the following three individual components:
\begin{description}
\item[Longitudinal detection range] The detection range must fulfil a defined safety requirement.
\item[Lateral detection accuracy] The lateral deviation of the detected lane must be accurate enough to enable a safe trajectory.
\item[Scene semantics] Depending on the road layout and velocity, a deviation in detection range or lateral detection accuracy can have different impacts.
\end{description}
\noindent
An overview of the individual components of the proposed safety metric and their composition to a single safety score is shown in Fig.~\ref{fig:overview}.

If only one lane marking can be detected, the detection will obtain a safety score of 0.00 since no safe planning is possible without a complete perception of the lane due to the uncertainty about the lane width. For edge cases like roads without a center lane marking, a calculated centerline of the two outer road bounds could be considered as second lane marking.

The proposed metric can be applied to single-frame lane detection as well as to a maintained detection over multiple frames.

\subsection{Longitudinal Detection Area}
\label{subsec:detection_area}

For safe lane detection, it is important to not only accurately detect the position of lane markings, but also to achieve a detection range that is sufficient to ensure safety.
In order to assess safety in the longitudinal direction, a criticality area should be defined that must be covered by the detection range. As introduced in Sec.~\ref{sec:related_work} there exist multiple approaches such as the RSS model~\cite{RSS} which define safety distances; however, they are defined for inter-vehicles interaction or for parking scenarios and thus cannot be applied to our metric. 
Possible approaches for the lane detection could be the area required for a lane change or emergency braking to stop before the end of a lane. Since a lane change is not possible in all situations we use the distance to perform an emergency braking maneuver such that the vehicle can at least stop if the lane ends. Moreover, an additional safety distance of \SI{10}{\percent} is included possible extensions of the braking maneuver due to environmental influences.

The resulting longitudinal safety distance $d_{long}$ is defined as shown in Eq.\eqref{eq:d_long}. $v_0$ describes the initial velocity of the ego vehicle, $a$ is the braking acceleration and $t_{delay}$ describes the processing delay until the braking process starts.

\begin{equation}
        \label{eq:d_long}
        d_{long} = 1.1\cdot (v_0 \cdot t_{delay} + \frac{v_0^2}{2a})
    \end{equation}
To determine the achieved detection range $d_{det}$ we use the minimum length of the two detected lane boundaries.  
If $d_{det}$ of the lane detection system exceeds $d_{long}$, a safe state for the longitudinal case is present as we would be able to perform an emergency braking or a lane change as maneuver to avoid a potential collision or a leaving of the roadway. This leads to a longitudinal safety score $s_{long}=1.00$. 

Otherwise, in the worst case the missing detection could lead to a collision if e.g. the lane ends.
Therefore, we perform an evaluation of a potential collision using the remaining velocity $v_{r}$ after braking for $d_{det}$ meters as shown in Eq.~\eqref{eq:v_remaining}.

\begin{equation}
        \label{eq:v_remaining}
        v_r = \sqrt{v_0^2 - 2\cdot a \cdot d_{det}} 
\end{equation}

To evaluate the potential collision and calculate the longitudinal safety score $s_{long}$, we consider $v_r$ to approximate the severity. A vehicle should stay on its own lane if a lane change is not possible; hence, we do not consider further semantics such as the direction and speed limit of the lane and adjacent lanes in this case. We use four severity levels to comply with the five level based safety score classification defined by Volk et al.~\cite{volk2020safety}. The fifth level stands for no collision and is therefore not listed. To categorize the velocities into severity levels, we considered different studies~\cite{FREDRIKSSON20101672, EffectsVehicleImpact} that investigated fatality rates and severity of accidents in relation to the impact velocities, road user classes, and averaged over different ages of affected individuals. Here, it is necessary to use results from such studies as it is not possible to determine the exact severity of a potential collision due to the high number of influencing factors. The resulting classification divided into vehicles and vulnerable road users (VRUs) is shown in Tab.~\ref{tab:crash_rating}. The velocities are resulting impact velocities including the impact angle and absolute velocities of the affected road users. The safety score is scaled linearly within the severity class, decreasing for higher velocities.

\begin{table}[h!]
    \centering
    \caption{Safety Score in relation to resulting impact velocities.}
    \begin{tabular}{c    r    r} \toprule
    Safety Score & Vehicles & VRUs \\ \midrule
    0.8 - 0.6  &$<$ \SI{8.3}{\metre\per\second}  &$<$ \SI{3.0}{\metre\per\second}  \\
    0.6 - 0.4  & \SIrange{8.4}{13.9}{\metre\per\second} &\SIrange{3.1}{8.3}{\metre\per\second}  \\
    0.4 - 0.2  & \SIrange{14.0}{16.7}{\metre\per\second} &\SIrange{8.4}{11.1}{\metre\per\second}  \\
    0.0  & $> $\SI{16.7}{\metre\per\second} &$> $\SI{11.1}{\metre\per\second}  \\ \bottomrule
    \end{tabular}
    \label{tab:crash_rating}
    \vspace*{-0.2cm}
\end{table}

\subsection{Lateral Detection Accuracy}
\label{subsec:lateral_deviation}
\begin{figure}[t]
    \centering
    \includegraphics[width=0.95\linewidth, page=3, trim= 4.5cm 5cm 4.5cm 5cm, clip]{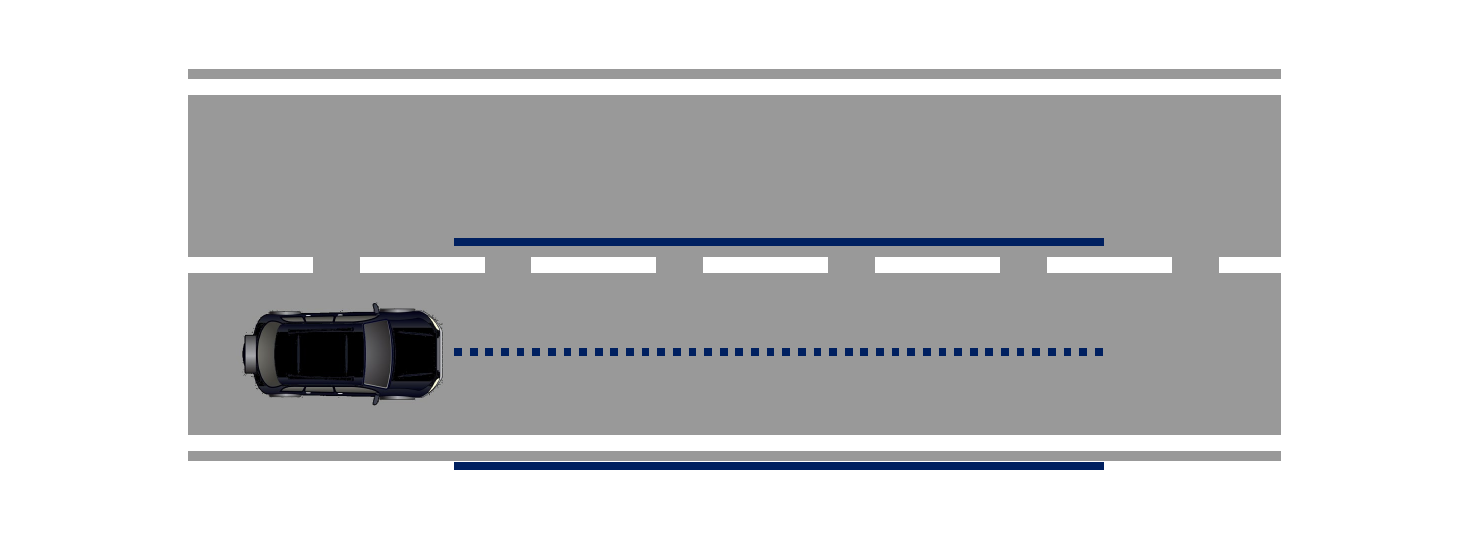}
    \caption{Scenario to demonstrate the advantages of the lateral evaluation based on the centerline. The solid blue lines illustrate the detected lane, dotted blue line symbolizes centerline. Both detected lines are inaccurate; however, following the centerline leads to a safe lateral behavior.}
    \label{fig:lateral}
\end{figure}

In addition to the longitudinal detection range for safe maneuvers, it is also necessary to take the lateral detection accuracy into account. The lane detection must be precise enough to plan a trajectory based on the detection without leaving the lane. Hence, the lateral detection accuracy is considered as second criteria for the safety evaluation. For the metric and the evaluation of the lateral detection accuracy we consider the deviation of the centerline between detected and the ground truth (GT) lane. The centerline describes the center of the detected or GT lane respectively and is taken into account since using the error of a single lane marking could falsify the safety score for some scenarios as shown in Fig.~\ref{fig:lateral}. An offset of lane markings detected on both sides of the lane, with a similar but limited size, can be considered safe as long as the vehicle remains within the lane. The evaluation of the lateral detection accuracy leads to the lateral safety score $s_{lat}\in[0,1]$.

Roads are planned to incorporate lateral range of movement between left to right lane boundary; for German roads regulations prescribe the movement tolerances shown in Tab.~\ref{tab:d_lat}. These values are based on the maximal allowed vehicle width of \SI{2.55}{\metre}.

\begin{table}[h!]
    \centering
    \caption{Lateral movement tolerance from~\cite{richter2010strassen}.}
    \begin{tabular}{c    c} \toprule
    Road type & Tolerance in [\si{\metre}] \\ \midrule
    Urban  & \SI{0.70}{\metre} \\
    Rural  & \SI{0.95}{\metre} \\
    Motorway & \SI{1.20}{\metre}  \\ \bottomrule
    \end{tabular}
    \label{tab:d_lat}
\end{table}

The real tolerable lateral deviation $th_{lat}$ between detected centerline and GT centerline depends on the width of the lane $w_l$ and the width of the vehicle $w_v$. It can be determined using Eq.~\eqref{eq:th_lat};
\begin{equation}
    \label{eq:th_lat}
    th_{lat} = \frac{w_l - w_v}{2} \pm x_{lat},
\end{equation}

where $x_{lat}$ describes an additional parameter which can be adapted based on the system design. If the vehicle should not be oriented on the centerline, the desired offset can be included with $x_{lat}$. In this case, it must be distinguished between a lateral deviation towards or against the offset direction. For our evaluation, we consider as the safest state when vehicle is positioned in the center of the lane; hence, $x_{lat}=\SI{0.0}{\metre}$ applies. 

The lateral deviation for safety purposes, denoted as $d_{lat}\in[0,\inf)$, is determined using Eq.~\eqref{eq:d_lat} where $CP_L$ describes a point on the centerline of the detected lane $L$ and $CP_{GT}$ the closest point on the GT centerline. Here, points are 3D positions in world coordinates.

\begin{equation}
    \label{eq:d_lat}
    d_{lat} = \lVert CP_L-CP_{GT} \rVert
\end{equation}

While $d_{lat}\leq th_{lat}$ holds, a vehicle following the centerline of the detected lane would stay on the lane. However, to maintain a safe status not the complete available range of movement should be used. Therefore, we set a limit to \SI{80}{\percent} of $th_{lat}$ which must not be undercut to incorporate a safety distance.
If $d_{lat}=0.0$, the vehicle drives exactly on the centerline which is considered as the safest status; hence, $s_{lat}=1.0$ must hold. Hence, for $d_{lat}\in [0.0, 0.8\cdot th_{lat}]$ a linear mapping of $s_{lat}\in[1.0, 0.8)$ is applied which stands for a safe but not perfect perception.
For $d_{lat} > 0.8\cdot th_{lat}$ the safety distance to the lane boundary is undercut and the vehicle following this lane is about to leave the lane. Hence, the lowest lateral safety score of $s_{lat}=0.8$ applies and the evaluation of scenario semantics (see Sec.~\ref{subsec:collision_relevance}) is conducted. 

However, the metric should not be too strict to achieve a meaningful result that complies with reasonable driving maneuvers. Hence, a single false detected point of the lane must not lead to a low safety score since single outliers can be filtered out and would not necessarily affect the trajectory planning. Therefore, the lateral detection safety is only affected if the lateral detection deviation of the centerline is present for at least a consecutive distance of $d_{min}$ as defined in Eq.~\eqref{eq:d_min}.
\begin{equation}
    \label{eq:d_min}
    d_{min}=t_{delay}\cdot v_0
\end{equation}

\subsection{Scenario Semantics}
\label{subsec:collision_relevance}
In the case that the lateral detection accuracy is insufficient it could lead to situations where the vehicle is close to leaving the lane or in the worst case leaves the lane. Driving on the wrong lane could lead to safety-critical scenarios or collisions and therefore must lead to a low safety score. The lateral detection accuracy evaluation (see Sec.~\ref{subsec:lateral_deviation}) is conducted to check if the vehicle is about to leave the lane. If the tolerable lateral deviation is exceeded it is indicated by $s_{lat}=0.80$ (see Sec.~\ref{subsec:lateral_deviation}). In this case, a more comprehensive statement about safety can be made by conducting a further evaluation of the possible consequences and their severity. This results in a scenario-dependent safety score $s_{scen}\in[0.0,0.8]$.

For the evaluation of the scenario semantics we consider the driving direction of the safety-critical adjacent lane $L_{adj}$, which describes the lane which the vehicle erroneously enters due to a lateral deviation. $L_{adj}$ can be of types: \textit{same direction}, \textit{opposite direction}, \textit{VRUs} or \textit{NoLane} which describes the area next to a paved road. The \textit{VRUs} case includes bicycle lanes and sidewalks, which are both used by VRUs. Additionally, we incorporate the speed limit of $L_{adj}$ and the ego velocity to estimate the severity of a potential collision if there is an other object on $L_{adj}$. 
For the classification of $s_{scen}$ the severity levels in relation to the impact velocities as shown in Tab.~\ref{tab:crash_rating} are used. To determine $s_{scen}$ the impact velocity resulting from the ego velocity, the speed limit of $L_{adj}$ and the angle between the ego lane and $L_{adj}$ are taken into account. Additionally, to incorporate the vulnerability of different classes of road users the classification into severity levels is adapted to lower velocities for the \textit{VRUs} case (see Tab.~\ref{tab:crash_rating}) as here potential collisions would include vulnerable road users. The \textit{VRUs} column of Tab.~\ref{tab:crash_rating} applies if $L_{adj}=VRUs$, otherwise the vehicles column for the impact velocity applies.

\subsection{Final Safety Score}
\label{subsec:safety_score}
Finally, to achieve an easily comparable and meaningful safety assessment, the results of the intermediate metrics presented in Sec.~\ref{subsec:detection_area}-\ref{subsec:collision_relevance} must be combined into a single safety score $S$. A weighted sum may result in a safety score that indicates an acceptable safety level based on a single high intermediate metric score. However, it is important to maintain a rather strict safety score. Thus, we differentiate based on the safety score of the lateral deviation intermediate metric to define the final safety score $S$ as shown in Eq.~\eqref{eq:S}.

\begin{equation}
    \label{eq:S}
    S = \begin{cases}
        min(s_{long}, s_{lat}) & s_{lat} > 0.80 \\
        min(s_{long}, s_{scen}) & \, \text{else}
\end{cases}
\end{equation}

If $s_{lat}>0.80$ applies, the vehicle is inside the tolerated lateral deviation and can remain in its own lane based on the detection, and therefore $s_{scen}$ has no influence. To ensure that $s_{long}$ is not neglected in this case, we consider the minimum of $s_{lat}$ and $s_{long}$ as safety assessment, selecting the most safety-critical intermediate metric result.
If the lateral detection accuracy is insufficient to achieve safe trajectory planning, this results in $s_{lat}=0.80$. In this case, for the final safety assessment, we consider $s_{scen}$ instead of $s_{lat}$ due to the highly safety-critical situation. If the vehicle departs from the lane, we can determine the safety criticality by assessing the potential severity based on the velocity and type of the adjacent lanes, which is represented by $s_{scen}$. As in the first case, we include $s_{long}$ to not neglect the required longitudinal detection range and use the minimum between the two relevant intermediate metric to obtain a meaningful safety score.

Since $S$ is not a percentage value as state-of-the-art performance metrics, a classification is introduced to enable an easy interpretability. 
We use an adapted version of the safety metric score classification proposed by Volk et al.~\cite{volk2020safety} which is shown in Tab.~\ref{tab:safety_classification}. This classification enables statements about safety, potential damages, and the severity of situations resulting from insufficient perception.
\begin{table}[h!]
\centering
\caption{Adapted safety metric score classification from~\cite{volk2020safety}.}
\begin{tabular}{c    l} \toprule
    {$\textit{\textbf{S}} \boldsymbol{\in}$} & {$\textit{\textbf{Classification}}$} \\ \midrule
    \multicolumn{1}{r}{{[}0.0 - 0.2{]}}  & \textbf{insufficient}, high risk of fatality \\
    \multicolumn{1}{r}{{(}0.2 - 0.4{]}} & \textbf{very bad}, existing risk for serious violation \\
    \multicolumn{1}{r}{{(}0.4 - 0.6{]}}  & \textbf{bad}, low probability of minor injuries  \\
    \multicolumn{1}{r}{{(}0.6 - 0.8{]}} & \textbf{good}, low risk of bodywork damage    \\
    \multicolumn{1}{r}{{(}0.8 - 1.0{]}} & \textbf{very good}, high probability of safe status  \\ \bottomrule
\end{tabular}
\vspace{-1mm}
\label{tab:safety_classification}
\end{table}

\section{RESULTS}
\label{sec:results}
State-of-the-art benchmarks such as~\cite{TuSimple, CULane, Fritsch2013ITSC, LLamas} use pixel-based performance metrics for evaluation and the corresponding datasets do not provide an accurate ground truth map and lack safety-relevant information about the vehicle, such as the velocity or the dimension, which are required for safety evaluation.
Thus, we use the CARLA simulator~\cite{CARLA} for the evaluation of the proposed metric. The data processing is conducted using the RESIST framework~\cite{resist}. A generic lane sensor by Gamerdinger et al.~\cite{gamerdinger2023cold} is used to simulate different sensing ranges as well as errors on the detection such as noise or offsets for the vehicle-local lane detection in Sec.~\ref{subsec:result_testcases}. The Lanelet framework~\cite{lanelet2} is used to store and handle the perceived lane information as well as the GT map. Additionally, we evaluate the two state-of-the-art lane detectors UFLD~\cite{qin2020ultra} and LaneATT~\cite{tabelini2021cvpr}. For both detectors we used the pretrained networks provided by the TURoad Lanedet framework~\cite{TuRoadLaneDet}.

We show the advantages of our safety metric motorway (Motorway1-Motorway4), rural (Rural1-Rural3) and urban (Town1-Town3) scenarios with 100 frames (\SI{10}{\second}) each, in comparison to the well known metrics precision, recall, and F1 as introduced in Sec.~\ref{sec:related_work}.
For performance evaluation we evaluate the lane point-wise over both lane boundaries with a distance of $\SI{0.10}{\metre}$ between two consecutive points.
To classify between TP and FP, we use a distance threshold for a detected point of $\SI{0.10}{\metre}$. Undetected points are classified as FNs, using $d_{long}$ as the labeling distance. 

We perform a single frame detection evaluation; however, the proposed safety metric is also capable to evaluate a lane detected and maintained over multiple frames.

\subsection{Limitations of Performance Metrics}
\label{subsec:result_testcases}

For most scenarios there exists a correlation between performance and safety; however, for some cases performance metrics lack meaningfulness.
The necessity of advanced safety metrics for the evaluation of perception systems is shown using three scenarios ($C_1$-$C_3$) as shown in Tab.~\ref{tab:testcases}. 

\begin{table}[h!]
\centering
\caption{Showcases to demonstrate the necessity of safety evaluation. * represents a don't care.}
\begin{tabular}{c    rrrr} \toprule
    Testcase & Velocity & $d_{det} (left/right)$ & $d_{lat}$ & $L_{adj}$\\ \midrule
    $C_S$  & \SI{13.89}{\metre\per\second} &(\SI{40}{\metre}, \SI{40}{\metre}) & $0.1\cdot th_{lat}$ & * \\
    $C_1$  & \SI{27.78}{\metre\per\second} &(\SI{30}{\metre}, \SI{60}{\metre}) & $0.1\cdot th_{lat}$ & * \\
    $C_2$  & \SI{13.89}{\metre\per\second} &(\SI{40}{\metre}, \SI{40}{\metre}) & $> th_{lat}$  & \textit{VRUs} \\
    $C_3$  & \SI{13.89}{\metre\per\second} &(\SI{40}{\metre}, \SI{40}{\metre}) & $0.2\cdot th_{lat}$  & * \\\bottomrule

\end{tabular}
\label{tab:testcases}
\end{table}

\begin{figure}[h!
]
    \centering
    \includegraphics[width=\linewidth, trim= 3.5cm 0cm 3.5cm 1.5cm, clip]{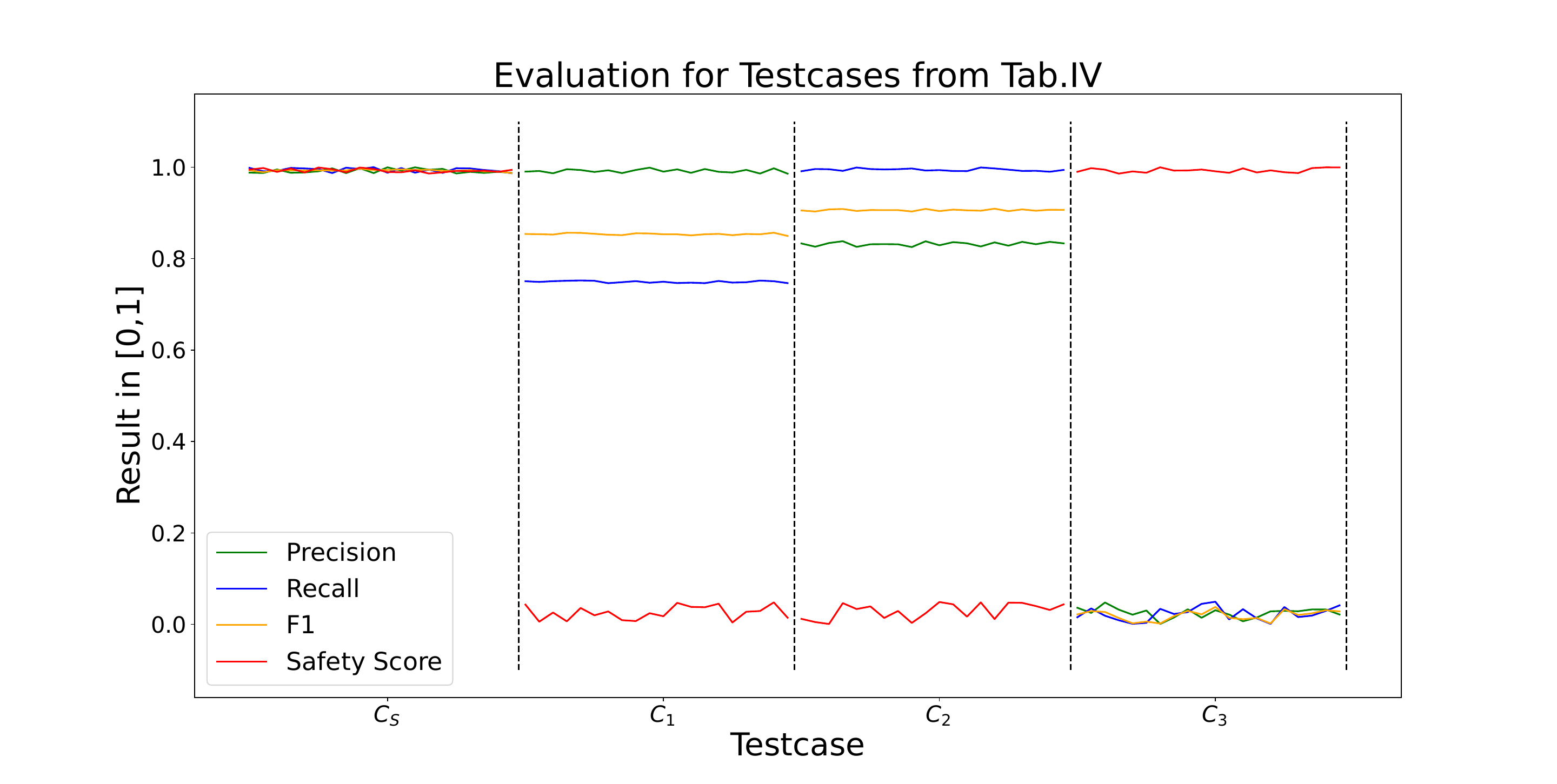}
    \caption{Performance and safety results for the testcases shown in Tab.~\ref{tab:testcases}. Vertical black lines divide between the testcases.} 
    \label{fig:result_showcase}
\end{figure}

Figure~\ref{fig:result_showcase} shows the safety score in comparison to precision, recall, and F1 for the testcases shown in Tab.~\ref{tab:testcases}.

\begin{table*}[t]
    \caption*{}
    \begin{minipage}{0.5\linewidth}
      \caption{UFLD Results, Safety Score ($\mu$, min, max).}
      \centering
        \begin{tabular}{l    rrrr} \toprule
    Scenario & Safety Score $S$ & Precision & Recall & F1\\ \midrule
    Motorway1& (0.93, 0.51, 0.99) & 0.4199 & 0.9810 & 0.5881 \\
    Motorway2& (0.90, 0.41, 0.99) & 0.4207 & 0.9943 & 0.5913\\
    Motorway3& (0.87, 0.00, 0.99) & 0.6028 & 0.6360 & 0.6189\\
    Motorway4& (0.39, 0.00, 0.92) & 0.5130 & 0.3203 & 0.3944\\
    Rural1& (0.47, 0.00, 0.99) & 0.3544 & 0.5472 & 0.4301\\
    Rural2& (0.49, 0.00, 0.93) & 0.0898 & 0.2287 & 0.1290\\
    Rural3& (0.47, 0.00, 0.99) & 0.1801 & 0.2981 & 0.2245\\
    Town1& (0.28, 0.00, 0.97) & 0.3549 & 0.5326 & 0.4260\\
    Town2& (0.39, 0.00, 0.98) & 0.2896 & 0.2145 & 0.2464\\
    Town3& (0.52, 0.00, 0.97) & 0.2263 & 0.4178 & 0.2936\\\bottomrule
\end{tabular}
\label{tab:results_UFLD}
    \end{minipage}%
    \begin{minipage}{0.5\linewidth}
      \centering
        \caption{LaneATT Results, Safety Score ($\mu$, min, max).}
        \begin{tabular}{l    rrrr} \toprule
    Scenario & Safety Score $S$& Precision & Recall & F1\\ \midrule
    Motorway1& (0.96, 0.89, 0.99) & 0.7838 & 1.0000 & 0.8788\\
    Motorway2& (0.97, 0.89, 0.99) & 0.6944 & 1.0000 & 0.8197\\
    Motorway3& (0.96, 0.90, 0.99) & 0.7365 & 0.9559 & 0.8320\\
    Motorway4& (0.56, 0.00, 0.93) & 0.6249 & 0.5442 & 0.5818\\
    Rural1& (0.62, 0.00, 0.98) & 0.4435 & 0.8025 & 0.5713\\
    Rural2& (0.67, 0.00, 0.97) & 0.4600 & 0.8627 & 0.6000\\
    Rural3& (0.13, 0.00, 0.98) & 0.4388 & 0.7702 & 0.5591\\
    Town1& (0.59, 0.00, 0.99) & 0.7269 & 0.8978 & 0.8034\\
    Town2& (0.39, 0.00, 0.99) & 0.6785 & 0.7109 & 0.6943\\
    Town3& (0.73, 0.00, 1.00) & 0.6989 & 0.9310 & 0.7984\\\bottomrule
\end{tabular}
\label{tab:results_LaneATT}
    \end{minipage} 
\end{table*}

For comparison we introduce a standard case $C_S$ where a vehicle is driving with $\SI{13.89}{\metre\per\second}$ with a sufficient detection range and a high lateral detection accuracy. The first case $C_1$ shows a vehicle with a high velocity and an insufficient detection range. The case $C_2$ is a vehicle with a lower velocity with a sufficient detection range but a partial lateral deviation of one lane border which leads to leaving of the lane.


For $C_S$ we can observe a safety score $S$ of about 1.00. This can be traced back on a sufficient $d_{det}$ which leads to $s_{long}=1.00$ and a highly accurate lateral detection accuracy resulting in $d_{lat}=1.00$.

In the case of $C_1$ the lateral detection accuracy is high; therefore $s_{lat}\approx 1.00$ applies. However, due to the high velocity, the detection range of $d_{det}=\SI{30}{\metre}$ of the left lane boundary is not sufficient to allow a safe maneuver. Based on a braking acceleration of $a=\SI{7.5}{\metre\per\second\squared}$, a velocity of $v_r=\SI{17.94}{\metre\per\second}$ remains during an emergency braking maneuver. Including the crash ratings introduced in Tab.~\ref{tab:crash_rating} this leads to $s_{long}=0.00$ which then leads to $S=0.00$. Based on the highly accurate lateral detection, we get a precision of about $\SI{100}{\percent}$. The recall is about \SI{75}{\percent}, because only for the left boundary \SI{30}{\metre} are missing and the right boundary is detected correctly. Both performance metrics show a good result even when the situation is very uncertain.

For $C_2$ the detection shows a sufficient detection range which leads to $s_{long}\approx 1.00$. However, a lateral deviation of the right lane boundary for about \SI{10}{\metre} could lead to an unwanted leaving from the lane to a sidewalk. Based on the velocity this leads to $s_{scen}=0.00$ and to $S=0.00$ as we have a safety critical situation in which VRUs could be harmed. As \SI{20}{\metre} out of \SI{30}{\metre} are correctly perceived and the left lane boundary is perceived correctly over \SI{30}{\metre}, this leads to a precision of about $\SI{83}{\percent}$ and a recall of about $\SI{100}{\percent}$ as the required detection range by Eq.~\eqref{eq:d_long} is only about \SI{16}{\metre}. These results indicate a good perception performance in an unsafe situation similar to $C_1$.

Test case $C_3$ shows the contrary behavior, where the safety metric also provides a more meaningful result compared to the performance metrics. $C_3$ represents the scenario shown in Fig.
~\ref{fig:lateral}. In this scenario, the lateral deviation would lead to a classification of the detected points as FP, and due to the lack of correct detection, the number of FNs is also high. Therefore, the precision and recall will be around 0.00. In terms of safety, a lateral deviation of $0.2\cdot th_{lat}$ would lead to $s_{lat}=0.95$. As for $C_S$ and $C_2$, the detection range is sufficient, resulting in $s_{long}\approx 1.00$. The resulting final safety score will be $S=0.95$, which represents a safe perception even if the performance metrics show low results.

\subsection{Evaluation of Lane Detectors}
\begin{figure*}[t]
    \centering
    \includegraphics[width=.95\linewidth, page=6, trim= 0cm 3cm 0cm 3cm, clip]{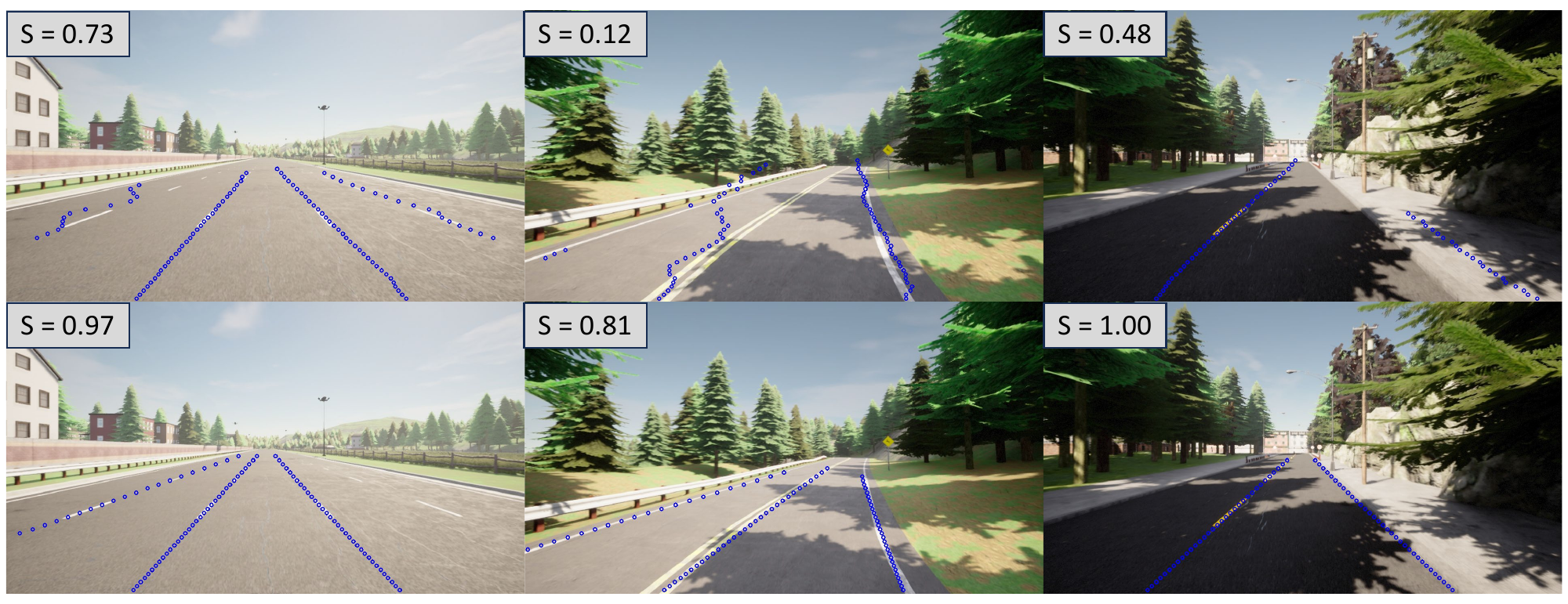}
    \caption{Extract of results with detected lanes by UFLD (top) and LaneATT (bottom) with the corresponding safety score of the frame for Motorway (left), Rural (mid) and Town (right).} 
    \label{fig:results}
    \vspace*{-0.4cm}
\end{figure*}

An overview of the obtained safety scores in comparison to precision, recall, and F1 is presented in Tab.~\ref{tab:results_UFLD} for UFLD and Tab.~\ref{tab:results_LaneATT} for LaneATT. The results for the safety score are calculated per frame and averaged over the scenario. Precision and recall are calculated over all frames. An extract of the safety metric results in combination with the detected lanes is presented in Fig.~\ref{fig:results}.

\subsubsection{UFLD}
When evaluating UFLD, we can observe similar results for the first three motorway scenarios.
They achieve an average safety score of about 0.90, which represents a high safety. Considering the recall of 0.63 to 0.99, there is a correlation with the high safety score; however, the precision indicates poor or unsafe detection. Based on the low precision, the F1 score of about 0.6 does not indicate a good performance either. This is based on a lateral deviation of about \SI{0.30}{\meter}, which leads to a high number of FPs and therefore affects the performance, while the lateral deviation is small enough to not negatively affect the safety.
For the fourth motorway scenario, the safety score and performance metrics indicate a poor perception. However, a safety score of 0.39 indicates a risk of serious injury, which is not indicated by the low performance.

In all rural scenarios, safety scores are approximately 0.48 with a minimum of 0.00 and a maximum of over 0.90. The safety score is 0.00 for undetected frames and about 1.00 when the lane is detectable. It can be observed that for a similar safety score, the performance metrics vary greatly. Therefore, performance metrics cannot be used with a threshold to determine safety.

For the town scenarios, UFLD achieves an average safety score ranging from 0.28 ( Town1) to 0.52 ( Town3). This can be attributed to the road architecture, where the right lane marker is the curb of the sidewalk, making detection significantly more difficult. Consequently, there is no detection in several frames, resulting in a minimum safety score of 0.00. Here, the F1 score indicates a better performance for Town1 with 0.42 than for Town3 with 0.29, which is in contrast to the safety scores obtained.

\subsubsection{LaneATT}
LaneATT achieves a safety score above 0.96 for three out of four scenarios. Considering the minimum safety, we can observe about 0.90 for these three scenarios, which indicates a safe state over the entire scenario. Only Motorway4 has a lower safety score of 0.56. The lower safety score is due to some frames with no detection, resulting in a safety score of 0.00.
We can observe high values for precision, recall and F1. However, considering that F1 is about 0.85, the performance seems to be limited even when the safety is high.
 
In Rural1 and Rural2, the safety score is about 0.65, which corresponds to a not negligible risk of collision. While precision indicates a poor performance, recall and F1 show a reasonable level of performance in an unsafe situation. This can also be observed for Rural3 with a mean safety of 0.13, indicating a very unsafe situation, which is not indicated by the performance metrics with values up to 0.77 for recall.

In the town scenarios, the detection system has difficulty identifying the lane with the curbstone as a lane marking, resulting in mean safety scores ranging from 0.39 (Town2) to 0.73 (Town3). The safety score per frame ranges from 0.00 up to approximately 1.00 for all scenarios, indicating that there are several frames without detection. However, when a detection is present, it is accurate and has a sufficient detection range. For all three scenarios, the precision is about 0.70, and the recall is up to 0.93 which indicates a good performance while the LSM indicates unsafe situations with a risk for up to serious injuries.

\subsection{Discussion}
As shown in the testcases in Sec.~\ref{subsec:result_testcases}, different relations between safety and performance metrics can exist. A better perception leads to higher performance and safety scores; however, a better performance does not necessarily correspond to a higher safety. 
Testcases $C_1$ and $C_2$ show that performance metrics are not suitable to evaluate safety. The F1 score as harmonic mean of precision and recall as well as the lower score of both with about 0.80 indicate a good performance, whilst the perception is not good enough to enable a safe status. A contrary behavior can be observed for $C_3$, here we can show that scenarios exist in which the performance metric show low scores whilst the perception is precise and long enough to be safe. 

Similar observations can be made when evaluating UFLD and LaneATT.
For scenarios like Town1 (LaneATT) with a recall of about 0.90 and a precision above 0.70, the safety can be 0.00 for some frames and even the mean of 0.59 indicates that a collision with minor injuries can occur.
In some cases, the safety metric indicates a safe perception while the precision is rather low. This results in an average lateral error of \SIrange{0.20}{0.40}{\metre}, which leads to a classification as FP for most points. However, the lateral detection accuracy is sufficient in terms of safety. Therefore, $S$ indicates a high safety, while the performance in terms of precision is rather low. 

This contradictory behavior, combined with the misclassification of the performance metrics in $C_1$ and $C_2$ as well as in the evaluation of the lane detectors, shows that none of the performance metrics is suitable for assessing the safety of lane detection.

\addtolength{\textheight}{-1cm}
\section{CONCLUSION \& OUTLOOK}
\label{sec:conclusion}

In this paper we proposed a novel and highly variable metric for assessing the safety of lane detection systems in autonomous driving. While for most scenarios performance and safety correlates, we have shown that there exist scenarios in which performance metrics are not suitable to evaluate lane detection and can indicate misleading results. Unlike these performance metrics such as precision, recall, and F1, our safety metric allows for a more comprehensive and meaningful evaluation in terms of safety.

The safety metric provides a methodology for an offline safety evaluation of lane detectors for a specific vehicle. Therefore, we determine the necessary longitudinal detection range using an adapted method from the well-known RSS model. The actual detection range is assessed in relation to the required safe detection range, taking into account various factors such as the ego velocity and the possible braking acceleration. Additionally, the system evaluates the lateral deviation of the lane detection. If this deviation poses a safety risk, the system takes into account the scene semantics and considers the hypothetical effect of a collision caused by the deviation of the lane corridor as an additional safety factor. A key advantage is that the proposed lane safety metric results in one single score between 0 and 1 which allows an easy comparability between different methods. Moreover, the categorization into five safety stages (insufficient, very bad, bad, good and very good) allows a fast safety assessment and simplifies the interpretation of the result. Additionally, the metric can be easily adapted to different motion controllers and road layouts and can be used for single-frame detection as well as with a lane management system over multiple frames.

It is important to note that this metric cannot guarantee the safety of an autonomous vehicle, as this metric only covers the detection of lanes but considerably extends existing object-based safety metrics. Thus, the metric provides an important component towards a comprehensive and composable safety assessment for environment perception and motion planning. 
The metric allows for an offline safety evaluation of a specific task in the pipeline for autonomous vehicles which is crucial for future tasks such as planning and contributes to the development of safe autonomous vehicles.

For future research, we plan to enhance our metric by incorporating additional scenarios and extend the metric to evaluate multi-lane detection and lane line segmentation. Additionally, we aim to evaluate various state-of-the-art lane detection methods under diverse environmental conditions to assess their safety. We aim to release the code to make it open to use for research to increase safety in autonomous driving.







\bibliographystyle{IEEEtran} 
\bibliography{literature.bib}

\end{document}